\documentclass[sigconf,natbib=true,anonymous=false,nonacm=true]{acmart}

\usepackage{subcaption}
\usepackage{hyperref}
\usepackage{url}
\usepackage{booktabs}
\usepackage{array}
\usepackage{xcolor}
\usepackage{rotating}
\usepackage{multirow}
\usepackage{tcolorbox}
\extrafloats{60}
\newif\ifcompact
\compacttrue
\AtBeginDocument{%
  }

\setcopyright{none}
\copyrightyear{2026}
\acmYear{2026}

\begin{document}

\title{Diagnosing Search Behavior and Failure Modes in Long-Horizon Search Agents}

\author{Qi Liu}
\email{qiliu6777@gmail.com}
\affiliation{%
  \institution{Renmin University of China}
  \city{Beijing}
  \country{China}
}

\author{Jiaxin Mao}
\email{maojiaxin@gmail.com}
\affiliation{%
  \institution{Renmin University of China}
  \city{Beijing}
  \country{China}
}

\author{Fengbin Zhu}
\email{fengbin@nus.edu.sg}
\affiliation{%
  \institution{National University of Singapore}
  \country{Singapore}
}

\author{Tat-Seng Chua}
\email{dcscts@nus.edu.sg}
\affiliation{%
  \institution{National University of Singapore}
  \country{Singapore}
}

\renewcommand{\shortauthors}{Liu et al.}

\begin{abstract}
Deep search agents answer difficult information-seeking questions by iteratively issuing search queries to gather supporting evidence, but it remains unclear whether and how greater search effort leads to better answers.
We study these questions through a trajectory-level diagnosis of long-horizon search agents. Using human-annotated document-level relevance judgments, we evaluate the evidence retrieved at each search step and separate two stages of agent behavior: what evidence an agent retrieves and how effectively it uses that evidence. This distinction further allows us to decompose failures into retrieval gaps, where the necessary evidence is never found, and utilization gaps, where relevant evidence is retrieved but not used correctly.
With the retrieval model and evaluation harness held fixed, we compare six agents on BrowseComp-Plus and further validate our findings on BrowseComp with an open-web search API. Across settings, we find that search effort and answer quality are only weakly aligned. Answer accuracy is better correlated with the quality of retrieved evidence, especially cumulative retrieval recall, than with the number of searches or the amount of context consumed. Useful evidence often appears early in the trajectory, yet agents tend to continue searching, producing a long tail of low-yield retrieval steps. At the query level, exploratory reformulations remain useful, but the best-performing agents issue far fewer redundant queries.

Overall, by systematically characterizing the search behavior and failure modes of long-horizon search agents, this work points to practical directions for building better deep research systems, including stronger query formulation, more effective evidence selection and context management, and stopping criteria based on whether sufficient supporting evidence has been retrieved.\footnote{Code and data at \url{https://github.com/liuqi6777/search_agent}}
\end{abstract}



\maketitle

\section{Introduction}

Deep search agents, such as OpenAI Deep Research and Tongyi DeepResearch \citep{Team2025TongyiDeepResearchTechnical}, are LLM-based systems for long-horizon information seeking and answering complex and multi-hop questions, where evidence is scattered across documents and must be searched for, checked, and synthesized before a conclusion holds \citep{Shi2025DeepResearchA}. The agent reasons, searches, reads what comes back, and decides turn by turn what to look for next, until it can commit to an answer \citep{Team2025MiroThinkerPushingthe}. Over one question, an agent may sustain retrieval and reasoning over tens of steps, and this long-horizon capacity is what separates it from single-shot retrieval-augmented generation.

However, evaluation practice has not caught up with these advances. Progress is still reported primarily through final-answer metrics, such as answer accuracy on BrowseComp \citep{Wei2025BrowseCompASimple}, while the trajectory that produces the answer remains an opaque byproduct. 

This poses concrete challenges for diagnosing agent behavior and developing improvement strategies:
\textbf{1) Wasted effort is mistaken for capability.} Long horizons and heavy tool use are routinely showcased as strengths, yet an effective and efficient search agent should instead adapt its search effort to what the problem genuinely requires: searching more only when the problem calls for it, not as a default behaviour.
\textbf{2) Bottlenecks get misdiagnosed.} A wrong answer may read as ``the agent could not find the evidence,'' so the fix is more or better search. Yet some errors occur with the decisive evidence already in hand, and more search cannot repair them. Differentiating between error types is essential for prescribing the right fix. 
\textbf{3) Persistence and floundering look the same.} Trajectories often run long after new evidence stops arriving, yet the score cannot tell late turns that are closing in on an answer from those that never anchor on the right direction, the very distinction between needing to search \emph{deeper} and needing to search \emph{better}.

To bridge this gap, we propose a diagnostic framework that opens up the trajectory itself. 
By grounding every step in human-annotated, document-level relevance, it measures retrieval-side and reasoning-side behaviour directly, and analyzes search agents from coarse to fine granularity.
Specifically, we organize our analysis into three key perspectives:
\textbf{1) Outcome attribution}: which signal actually tracks final accuracy: search effort or gold-evidence recall?
\textbf{2) Failure-mode taxonomy}: every error is sorted into a \emph{retrieval gap} (supporting evidence never found) or a \emph{utilization gap} (evidence found but still incorrect).
\textbf{3) Trajectory mechanism and query strategy}: each trajectory is followed episode by episode, tracking how the agent searches and when and where evidence arrives.

We conduct our analysis with six agents, spanning two scale tiers: three mid-scale (Tongyi DeepResearch \citep{Team2025TongyiDeepResearchTechnical}, Qwen3.5-35B-A3B \citep{qwen2025qwen3}, gpt-oss-120b \citep{openai2025gptoss}) and three frontier-scale (Kimi K2.6 \citep{kimi2026k26}, GLM 5.1 \citep{zeng2026glm}, Deepseek V4 Pro \citep{deepseekai2026deepseekv4}). All are run under the same ReAct framework, tools, and retriever on BrowseComp-Plus \citep{Chen2025BrowseComp-PlusAMore}, whose fixed corpus and human-annotated qrels supply the relevance grounding; we further validate the main conclusions on BrowseComp \citep{Wei2025BrowseCompASimple} with open-web search.

Our analysis distills into five findings, which trace a single thread from symptom through mechanism to prescription: 

\textbf{1) More effort does not buy more accuracy.} Across the six agents we analyze, neither search volume nor context spend predicts accuracy; if anything, the more accurate agents are also the cheaper ones (§\ref{sec:results-attribution}).

\textbf{2) What does track accuracy is retrieval recall.} Accuracy follows how much of the gold evidence an agent's queries cumulatively retrieve; on the same question, the agent that retrieves the gold evidence answers correctly most of the time, while the agent that misses it almost never does (§\ref{sec:results-attribution}, §\ref{sec:results-synthesis}).

\textbf{3) Failures split into two gaps with opposite fixes.} When agents do fail, the error is either a \emph{retrieval gap} --- the evidence was never surfaced --- or a \emph{utilization gap} --- it was surfaced, yet the answer still went wrong. Agents at nearly identical accuracy can sit at opposite ends of this balance, so the fix is per-diagnosis rather than per-leaderboard: better queries repair the former, better verification the latter (§\ref{sec:results-failure}).

\textbf{4) Evidence early or never, then a wasted tail.} The great majority of search episodes add no new evidence, and the waste is not spread evenly: whatever evidence a trajectory will surface, it surfaces in the early episodes. Correct runs assemble their evidence early and stop soon after; incorrect runs surface only part of it, or none at all, yet keep searching far longer without recovering the rest --- their extra length is a wasted tail, not late progress (§\ref{sec:results-mechanism}).

\textbf{5) Stronger agents search cleaner, not more.} The behavioral signature of strength is query discipline: redundant re-queries are rare but mark failure. Query strategy is not decisive: agents favor different styles, from broad pivoting to anchor-and-verify, and what counts is whether queries \emph{land} on evidence (§\ref{sec:results-query}).

Together, the findings argue that the next gains lie not in deeper search but in better-directed search (§\ref{sec:discussion}):
\textbf{1) For weaker agents, fix the queries:} improve query direction and search strategy, so that the agent recognizes when a hypothesis has failed and pivots to a new angle, instead of re-asking near-paraphrases of queries that already failed.
\textbf{2) For stronger agents, shift effort to verification:} the strongest agent already sits near the oracle-reader ceiling; the headroom that remains lies in verification and answer normalization, not in more retrieval.
\textbf{3) For both, improve at the harness level:} snippet-stream management so old results do not dominate the context window, visit gating against low-value page opens, and evidence-driven stopping in place of fixed budgets.

The paper contributes (i) a diagnostic framework for search agents grounded in relevance; (ii) a controlled comprehensive study of six agents and the five findings above, tracing search behavior and failure modes from coarse to fine; and (iii) a planned release of trajectories and code to support follow-up analysis.

\section{Related Work}
\label{sec:related}

\paragraph{Deep research agents and their evaluation.} Deep research agents extend ReAct \citep{yao2023react} to long-horizon information seeking by interleaving search, document fetching, and synthesis \citep{Xi2025ASurveyof, Shi2025DeepResearchA}; recent systems include proprietary \citep{kimi2026k26} and open variants \citep{Team2025TongyiDeepResearchTechnical, Du2026OpenSeekerDemocratizingFrontier, Li2026OpenResearcherAFully, Li2025WebSailorNavigatingSuper-human} alongside search-enhanced reasoning frameworks \citep{Li2025SearchO1Agentic, Li2025WebThinkerEmpoweringLarge}, with a parallel training literature on trajectory synthesis, RL for search-augmented reasoning, and tool-call efficiency \citep{Jin2025Search-R1TrainingLLMs, Chen2025ReSearchLearningto, Wang2025OTCOptimalTool}. Progress is reported as end-to-end accuracy on BrowseComp \citep{Wei2025BrowseCompASimple}. Our work is orthogonal: rather than building or evaluating an agent end-to-end, we open the trajectory and ask which steps pay off.

\paragraph{Analyzing agent search behavior.} Recent works examine what happens \emph{during} search rather than only at its endpoint. \citet{Ning2026AgenticSearchin} characterize $14$M$+$ requests from DeepResearchGym \citep{Coelho2025DeepResearchGymAFree} at log scale but without per-query relevance supervision; \citet{Chen2025BrowseComp-PlusAMore} provide the human-annotated qrels we build on, but use them to compare retrieval backends rather than diagnose the agent's process; AgentIR \citep{Chen2026AgentIRReasoning-AwareRetrieval} shows reasoning traces are themselves retrieval signals, motivating the snippet-first reading our findings support; other process-level work studies confidence / stopping calibration and trajectory-level evaluation \citep{Ou2025BrowseConfConfidence-GuidedTest-Time, Sun2026DeepSearchwith, Chen2026TRACETrajectory-AwareComprehensive}, and reasoning--action pathologies in which agents over-reason relative to acting \citep{Cuadron2025DangerOverthinking}. None combines the three properties our diagnosis requires --- \emph{per-query document-level qrels}, \emph{multiple agents}, and a \emph{single shared harness and retriever} --- the conjunction that makes the retrieval-vs.-utilization attribution \emph{deterministic} (decided by qrels rather than an LLM judge) and comparable across agents.



\section{Setup}

\subsection{Agent Settings}

\paragraph{ReAct paradigm.} We focus on agents that follow a ReAct-style \citep{yao2023react} search-and-reason loop. Given a question $q$, an agent's policy $\pi_\theta$ produces, at each step $t$, a reasoning trace $\tau_t$ and an action $a_t$ conditioned on the running history $h_t = (q, \tau_1, a_1, o_1, \ldots, \tau_{t-1}, a_{t-1}, o_{t-1})$: $(\tau_t, a_t) \sim \pi_\theta(\cdot \mid h_t)$.
Each action is either a tool call or a terminal answer; non-terminal actions yield an observation $o_t = \mathcal{T}(a_t)$ from the environment $\mathcal{T}$, which is appended to the history. A \emph{trajectory} is the resulting finite sequence $\xi = (\tau_1, a_1, o_1, \ldots, \tau_T, a_T)$, ending either with a terminal answer or with an early termination due to a step cap, context overflow, or tool error.

\paragraph{Tool interface.} The tool interface contains two tools, \texttt{search} and \texttt{visit}.
Both tools are routed to a local backend over the official BrowseComp-Plus corpus. The \texttt{search} tool issues a query $q'$ and returns the top-$K{=}5$ results retrieved by Qwen3-Embedding-8B \citep{zhang2025qwen3embedding}. Each result contains the document id, retrieval score, and a snippet truncated to $512$ tokens. The \texttt{visit} tool fetches the full document text given a document id.

\begin{table*}[t]
\centering
\small
\caption{Overall performance and per-question effort/cost on BrowseComp-Plus. \emph{Acc}: accuracy; \emph{Gold Rec.}: mean gold-document recall; \emph{Inc.}: incomplete rate; \emph{Q/turn}: queries per search turn ($>1$ = batching); effort/cost columns are \emph{median (mean)}.}
\label{tab:overall}
\begin{tabular}{lccccccc}
\toprule
Agent & Acc (\%) & Gold Rec.\ (\%) & Inc. (\%) & Search & Visit & Ctx (K) & Q/turn \\
\midrule
gpt-oss-120b           & 38.0 & 52.0 & 12.8 & 33 (32) & 1 (1.8) & 86 (83) & 1.00 \\
Tongyi-DR              & 52.2 & 64.2 & 38.0 & 30 (28) & 3 (3.8) & 84 (77) & 1.00 \\
Qwen3.5-35B-A3B        & 55.2 & 63.0 &  1.3 & 19 (22) & 1 (1.6) & 66 (74) & 1.74 \\
\midrule
Deepseek V4 Pro        & 68.6 & 75.4 &  0.7 & 21 (29) & 2 (1.8) & 72 (89) & 1.49 \\
Kimi K2.6              & 69.2 & 78.2 & 16.5 & 22 (27) & 1 (1.2) & 68 (93) & 1.01 \\
GLM 5.1                & 74.1 & 78.7 &  9.8 & 14 (20) & 2 (1.6) & 52 (73) & 1.87 \\
\bottomrule
\end{tabular}
\end{table*}

\paragraph{Agents.} We evaluate six agents in two scale tiers. All six have open weights, so the study never mixes open and closed systems and every rollout is in principle reproducible; what separates the tiers is model scale, and with it how we serve them. The \emph{mid-scale} tier comprises Tongyi DeepResearch \citep{Team2025TongyiDeepResearchTechnical}, Qwen3.5-35B-A3B \citep{qwen2025qwen3}, and gpt-oss-120b \citep{openai2025gptoss}, none above roughly $120$B total parameters; we serve their released checkpoints locally. The \emph{frontier-scale} tier comprises Kimi K2.6 \citep{kimi2026k26}, GLM 5.1 \citep{zeng2026glm}, and Deepseek V4 Pro \citep{deepseekai2026deepseekv4}, each substantially larger and reached through its provider's API. All six run through the same ReAct harness, which exposes \texttt{search} and \texttt{visit} as native tools and records reasoning traces, tool calls, tool returns, and the final answer per question. Each rollout is capped at $128$ turns and $150$ minutes; hitting either cap without an answer is recorded as \emph{incomplete}, as analyzed in §\ref{sec:results-failure}. The models use their recommended sampling parameters.

\subsection{Benchmark and Evaluation Signals}
\label{sec:signals}
\label{sec:impl}

We instantiate our analysis on \textbf{BrowseComp-Plus} \citep{Chen2025BrowseComp-PlusAMore}, a depth-oriented information-seeking benchmark of $830$ English questions paired with a fixed corpus and human-annotated qrels. We adopt it as our sole benchmark because it is the only public benchmark in this regime that provides the per-query, document-level relevance judgments our framework requires. Holding the corpus, retriever interface, and supervision signal fixed across agents isolates behavioral differences from setup confounds. Over the fixed ${\sim}100$K-document corpus, each question has an average of $6.10$ evidence-graded documents and $2.90$ gold-graded documents; the gold set is a strict subset of the evidence set.

BrowseComp-Plus's qrels come in two layers: a broader \emph{evidence-qrels} set covering documents with topical relevance to the question, and a stricter \emph{gold-qrels} set covering documents whose content is sufficient to derive the gold answer. We use both, but for different diagnoses. In the cross-agent attribution and episode-level mechanism analyses of §\ref{sec:results-attribution}--\ref{sec:results-mechanism}, $E(q)$ denotes the evidence-qrels set; a retrieved document counts as evidence iff its corpus document id appears in $E(q)$, regardless of surface relatedness. For the failure-mode classification in §\ref{sec:results-failure}, we use the stricter gold-qrels set $G(q) \subseteq E(q)$, since calling a wrong answer a \emph{utilization} failure requires that the trajectory have surfaced evidence sufficient to derive the right answer.

Both qrel signals are deterministic and independent of LLM judging, so the retrieval-side metrics in §\ref{sec:results-mechanism} and the failure classification in §\ref{sec:results-failure} are not confounded with answer-judging noise. End-to-end answer correctness is the only judgment task in our pipeline. Following the official BrowseComp protocol, GPT-4o judges each prediction against the BrowseComp-Plus gold answer with binary correct/incorrect labels; we use \texttt{gpt-4o-2024-08-06} at temperature $0$. End-to-end correctness therefore rests on this single judge, while the retrieval-side metrics do not, localizing any judge noise to the accuracy axis alone.

\section{Diagnostic Analysis}
\label{sec:analysis}

We layer the analysis from coarse to fine. §\ref{sec:results-overview} first profiles what each agent achieves and at what cost; the rest of the section then answers six research questions:

\noindent\textbullet\ \textbf{RQ1:} Which aggregate signal tracks accuracy --- search volume, context spend, or retrieval recall (§\ref{sec:results-attribution})?

\noindent\textbullet\ \textbf{RQ2:} When an agent fails, did it never surface the evidence, or surface it and still answer wrong (§\ref{sec:results-failure})?

\noindent\textbullet\ \textbf{RQ3:} Inside a trajectory, when does evidence arrive and what do the remaining turns contribute (§\ref{sec:results-mechanism})?

\noindent\textbullet\ \textbf{RQ4:} How do stronger agents search differently (§\ref{sec:results-query})?

\noindent\textbullet\ \textbf{RQ5:} Do these effects hold within the same question, with difficulty held fixed (§\ref{sec:results-synthesis})?

\noindent\textbullet\ \textbf{RQ6:} Do the findings generalize to open-web search (§\ref{sec:results-openweb})?

\subsection{Overall Performance and Effort Profile}
\label{sec:results-overview}

\emph{What does each agent achieve, and what does it cost?} Table~\ref{tab:overall} reports accuracy, mean gold recall, incomplete rate, and per-question effort: \texttt{search}/\texttt{visit} calls and final context. Accuracy ranges from gpt-oss-120b's $38.0\%$ to GLM 5.1's $74.1\%$; the frontier-scale tier, at $68.6\text{--}74.1\%$, sits above the mid-scale tier, at $38.0\text{--}55.2\%$, on both accuracy and mean gold recall. Effort does not follow accuracy. GLM 5.1 has the top accuracy with the lowest cost, a median of $14$ search calls and $52\,$K context, while gpt-oss-120b has the lowest accuracy and searches the most, with $33$ calls and $86\,$K context. Mean gold recall exceeds answer accuracy for every agent by $4.6\text{--}14.0$ points, so each agent surfaces more gold-qrels evidence than it converts into a correct answer. Tongyi-DR is the cost outlier: $38.0\%$ of its rollouts terminate without a final answer, overwhelmingly due to context-window overflow.

\paragraph{Per-turn query batching decouples search volume from turn count.} Agents differ markedly in how many queries they issue within a reasoning turn. In Table~\ref{tab:overall}, GLM 5.1, Qwen3.5-35B-A3B, and Deepseek V4 Pro batch $1.5\text{--}1.9$ queries as parallel tool calls, whereas gpt-oss-120b, Kimi K2.6, and Tongyi-DR issue essentially one. Search \emph{turns}, the round-trips that pace context growth, therefore spread more than search \emph{calls}: GLM 5.1 and gpt-oss-120b differ by only $1.6\times$ in total queries but by $3.0\times$ in turns, $10.6$ versus $31.8$. Much of the turn-count spread is batching rather than search depth, a turn-efficiency lever revisited in §\ref{sec:results-query}.

\begin{figure*}[t]
\centering
\includegraphics[width=\textwidth]{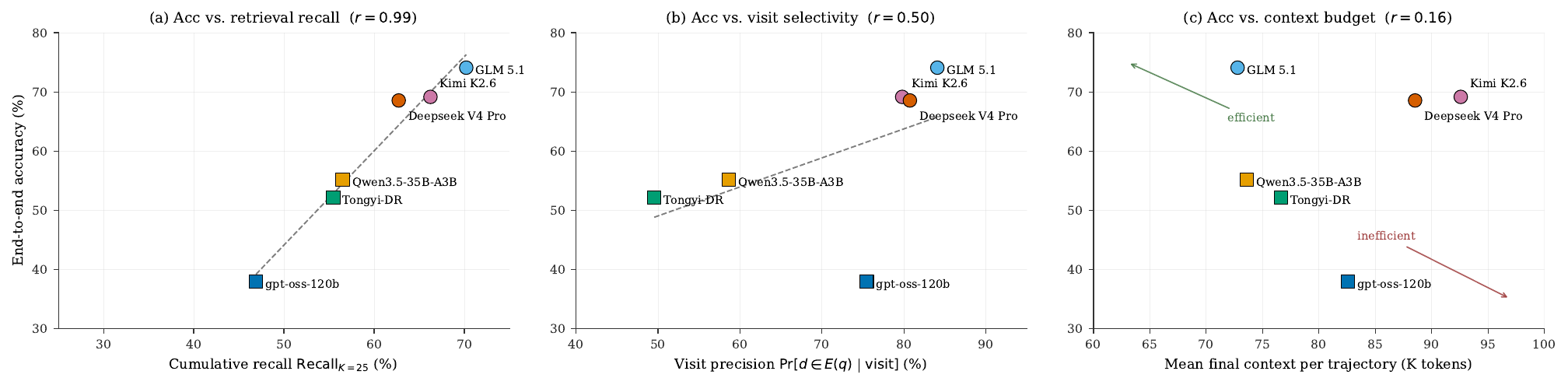}
\caption{Cross-agent attribution. End-to-end accuracy vs.\ \emph{(a)} padded cumulative recall at $K = 25$, \emph{(b)} visit precision, \emph{(c)} mean final context size. Pearson $r$ in panel titles. Stability (bootstrap CI): recall $[0.96,1.00]$, visit precision $[0.37,0.62]$, context $[-0.06,0.35]$.}
\label{fig:acc-vs-x}
\end{figure*}

\subsection{RQ1: Which Signal Tracks Accuracy?}
\label{sec:results-attribution}

\emph{When agents get a question right or wrong, which aggregate signals separate the two outcomes?} We compare three cross-agent candidates in Figure~\ref{fig:acc-vs-x} --- retrieval recall, visit precision, and context budget --- then ask how much search budget is needed before the relevant signal saturates.

\paragraph{Setup.} We group a trajectory $\xi$ into \emph{episodes} $k = 1,\dots,M$, where each episode starts with one \texttt{search} call and includes the \texttt{visit}s before the next search. For question $q$, let $E(q)$ be the evidence-qrels set and $G(q) \subseteq E(q)$ the stricter gold-qrels set (§\ref{sec:signals}). The retrieval trace is then summarized as
\begin{align*}
R_k(\xi) &\subseteq E(q) && \text{evidence surfaced in episode } k,\\
C_k(\xi) &= \bigcup_{j \le k} R_j(\xi) && \text{evidence surfaced up to episode } k,\\
\Delta_k(\xi) &= C_k(\xi) \setminus C_{k-1}(\xi) && \text{new evidence added by episode } k.
\end{align*}
The cross-agent recall in Figure~\ref{fig:acc-vs-x}a is \emph{padded cumulative recall}:
\[
\mathrm{Recall}_K = \frac{1}{N}\sum_q \frac{|C_{\min(K,M)}(\xi)|}{|E(q)|}.
\]
It measures evidence-qrels recall by episode $K$, holding shorter trajectories at their final value. This differs from Table~\ref{tab:overall}'s gold recall, which uses $G(q)$. For gold-specific events, define $G^*_k(\xi) = C_k(\xi) \cap G(q)$ and $G^*(\xi) = G^*_M(\xi)$; the first gold hit occurs at the smallest $k$ with $G^*_k(\xi) \neq \emptyset$. Finally, $\mathrm{success}(\xi,q)=1$ iff the trajectory terminates with $a_T=(\textsc{answer},\hat{y})$ and $\hat{y}$ matches the gold answer.

\paragraph{Accuracy tracks retrieval recall; no agent sits on the context-efficiency frontier.} The three aggregate metrics relate to accuracy very differently (Figure~\ref{fig:acc-vs-x}). Across agents, cumulative recall at $K = 25$ is the strongest correlate, with Pearson $r = 0.99$; visit precision is moderate at $r = 0.50$, with Qwen3.5-35B-A3B and Tongyi-DR lagging; final context size is essentially unrelated at $r = 0.16$. Since these are six-point correlations, we read them descriptively and rely on the within-question panel in §\ref{sec:results-synthesis} for inference. A paired bootstrap over the $830$ shared questions with $B = 10{,}000$, plus leave-one-agent-out, gives the intervals in Figure~\ref{fig:acc-vs-x}'s caption: recall stays at or above $0.96$ and exceeds context in every resample, context spans zero, and visit precision is fragile. Panel (c) makes the efficiency story concrete: only GLM 5.1 sits in the efficient upper-left, reaching $74.1\%$ accuracy at the lowest context budget on either statistic, $73\,$K mean and $52\,$K median, while gpt-oss-120b lands lower-right. Context efficiency is therefore an available lever, not one the strongest agents already optimize.

\begin{figure*}[t]
\centering
\includegraphics[width=\textwidth]{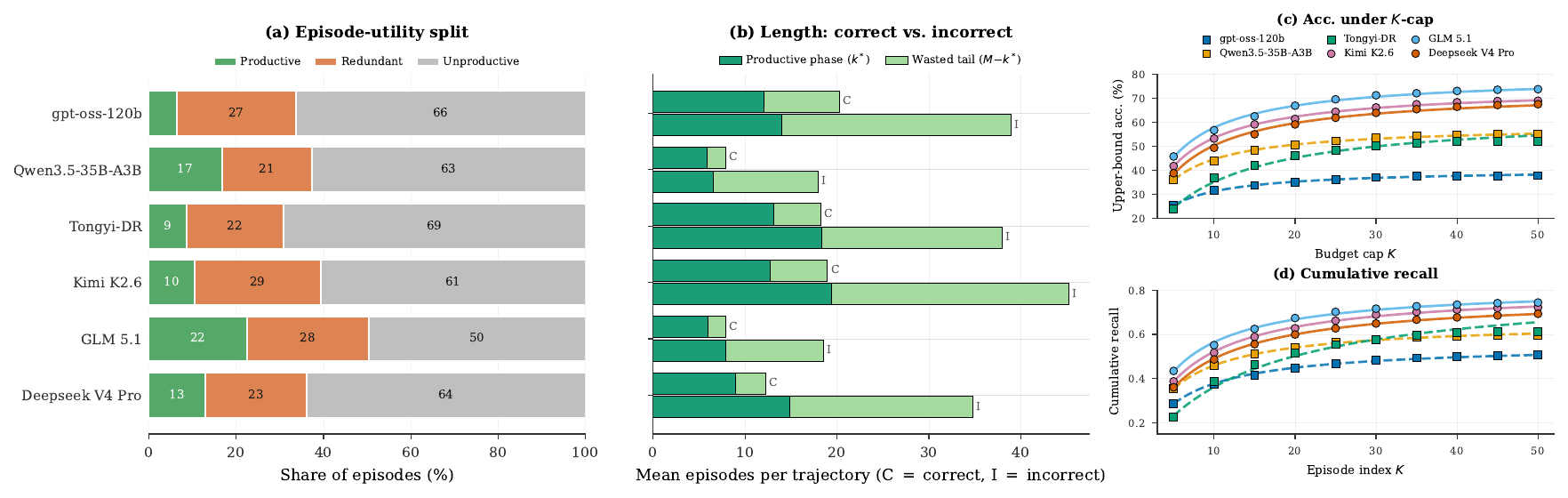}
\caption{Episode-level mechanism. \emph{(a)} Episode-utility split per agent; redundant$+$unproductive = no-new-evidence share ($77.5\text{--}93.6\%$). \emph{(b)} Trajectory length by correctness: productive phase (up to $k^*$) plus wasted tail; incorrect trajectories are $1.9\text{--}2.9\times$ longer yet saturate at similar $k^*$. \emph{(c,d)} Upper-bound accuracy $\widehat{\mathrm{Acc}}_K$ and padded cumulative recall vs.\ $K$.}
\label{fig:traj-combined}
\end{figure*}

\paragraph{A $K = 20$-episode cap retains up to $86\text{--}92\%$ of accuracy.} We estimate a budget-side upper bound by asking which successful full rollouts had already hit gold by episode $K$:
\[
\widehat{\mathrm{Acc}}_K =
\frac{1}{N}\sum_q
\mathbf{1}\{\mathrm{success}(\xi,q)=1 \ \wedge\  (G^*(\xi) = \emptyset \ \vee\ \exists k \le K: G^*_k(\xi) \neq \emptyset)\},
\]
with $N=830$.\footnote{This assumes the eventual answer is unchanged under early stopping; hard truncation may yield lower accuracy.} A question is therefore retainable if the full rollout answered correctly and either hit gold by episode $K$ or never hit gold at all, in which case the cap could not have cost it anything. As Figure~\ref{fig:traj-combined}c shows, $\widehat{\mathrm{Acc}}_K$ saturates as sharply as recall: $K = 20$ retains $86.3\text{--}92.4\%$ of each agent's accuracy, and $K = 40$ retains $97.0\text{--}99.8\%$. Since median rollouts use $14\text{--}33$ search calls in Table~\ref{tab:overall}, a $K = 20$ cap barely shortens the typical trajectory yet concedes at most $14\%$ of accuracy; §\ref{sec:results-mechanism} unpacks when gold actually lands within a trajectory.

\subsection{RQ2: Retrieval vs.\ Utilization Gap}
\label{sec:results-failure}
\label{sec:results-oracle}

\emph{When an agent is wrong, is it because it never surfaced the evidence or because it surfaced it and still answered wrong?} We call the first case a retrieval gap and the second a utilization gap, then analyze their distribution and intervention implications: retrieval-side fixes for RG, and a mix of reasoning and evaluation fixes for UG.

\begin{figure*}[t]
\centering
\includegraphics[width=\textwidth]{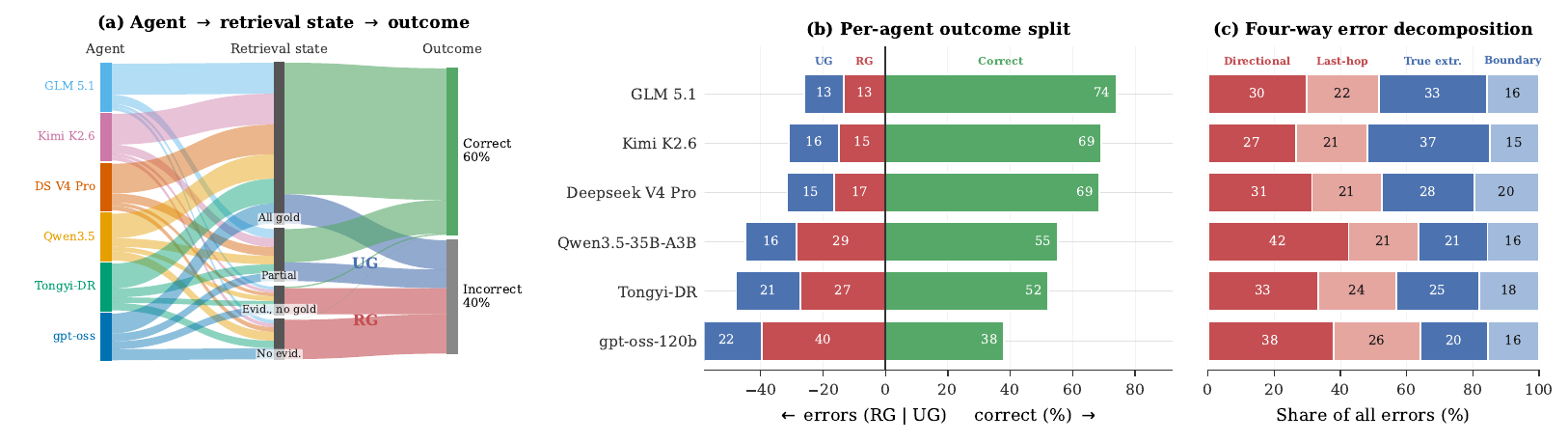}
\caption{Failure modes. \emph{(a)} Agent $\to$ retrieval state $\to$ outcome flow ($n = 4980$); the RG/UG split is \emph{which} retrieval state the incorrect flows leave. \emph{(b)} Per-agent outcome split (correct vs.\ RG $\mid$ UG), sorted by accuracy. \emph{(c)} Four-way error decomposition (directional / last-hop RG, true-extraction / boundary UG); warm = retrieval side, cool = utilization side.}
\label{fig:failure-combined}
\end{figure*}

\paragraph{Definitions.} Using $G^*(\xi)$ and $\mathrm{success}(\xi, q)$ from §\ref{sec:results-attribution}, we label each question:
\begin{equation*}
\begin{aligned}
&\textbf{correct:} && \mathrm{success}(\xi, q) = 1, \\
&\textbf{utilization gap:} && \mathrm{success}(\xi, q) = 0 \text{ and } G^*(\xi) \neq \emptyset, \\
&\textbf{retrieval gap:} && \mathrm{success}(\xi, q) = 0 \text{ and } G^*(\xi) = \emptyset.
\end{aligned}
\end{equation*}
Incompletes, where no $\hat{y}$ is emitted, are folded in by the same rule. RG points to query or retrieval-side fixes, while UG points to extraction or reasoning-side fixes. Figure~\ref{fig:failure-combined}a lays out the whole partition as a flow; panels b and c sharpen the per-agent split.

\paragraph{Retrieval gap dominates errors in five of six agents.} In Figure~\ref{fig:failure-combined}, RG accounts for $51.6\text{--}64.1\%$ of errors for five agents. Kimi K2.6 is the exception, with $52.0\%$ UG, consistent with its high mean gold recall of $78.2\%$. The balance is not pinned by accuracy: GLM 5.1 reaches $74.1\%$ accuracy and Tongyi-DR only $52.2\%$, yet their RG shares are within $6$\,pp, so very different agents can still call for similar retrieval-side interventions; conversely, Kimi K2.6 and Deepseek V4 Pro land within a point of each other in accuracy yet on opposite sides of the RG/UG split, so nearly identical leaderboard positions can call for opposite fixes. Restricting to answered-only trajectories sharpens two cases: Tongyi-DR's RG share falls from $56.9\%$ to $39.0\%$, because overflow incompletes had often already retrieved gold, and Kimi K2.6 falls from $48.0\%$ to $24.4\%$ RG.\footnote{Tongyi-DR's overflow subset is systematically harder: the other five agents score $10.5\text{--}59.0\%$ on it, well below their full-set figures, so Tongyi-DR's answered-only accuracy of $84.1\%$, against $52.2\%$ overall, partly reflects an easier residual.} §\ref{sec:results-mechanism} takes up the underlying post-gold redundancy.

\paragraph{RG and UG each split into two subtypes that prescribe different fixes.} Each gap splits by \emph{how far} the trajectory got through the four retrieval states in Figure~\ref{fig:failure-combined}a. A retrieval gap is \emph{directional} when no evidence document is reached, so the agent never landed in the right topical neighborhood; it is \emph{last-hop} when evidence is reached but the gold-graded subset is missed. The fixes differ: hypothesis re-anchoring for the first case, terminal-query precision for the second. Directional RG dominates everywhere, accounting for $55\text{--}67\%$ of each agent's RG in Figure~\ref{fig:failure-combined}c and peaking at $66.7\%$ for Qwen3.5-35B-A3B. A utilization gap is a \emph{true extraction failure} when all gold was retrieved but the answer is still wrong, and a \emph{boundary} UG when only part of the gold was retrieved. Kimi K2.6 is the most extraction-bound, with $71.4\%$ true UG; gpt-oss-120b's UG is the most boundary-heavy at $43.2\%$, so its nominal RG/UG split under-counts retrieval failure.

The subtype decomposition implies per-agent prescriptions: break initial-hypothesis anchoring for Qwen3.5-35B-A3B, which has the largest directional share; combine deeper terminal-step search with stronger answer integration for Kimi K2.6, which is jointly high on last-hop RG and true-extraction UG; and treat Deepseek V4 Pro as the most balanced case. Across agents, post-gold redundancy remains high: after the first gold hit, $57.9\text{--}74.0\%$ of later episodes add no new evidence. For Tongyi-DR, this wasted post-gold rate reaches $66.0\%$ and helps drive its $38.0\%$ overflow rate.

\begin{figure}[t]
\centering
\includegraphics[width=0.52\linewidth]{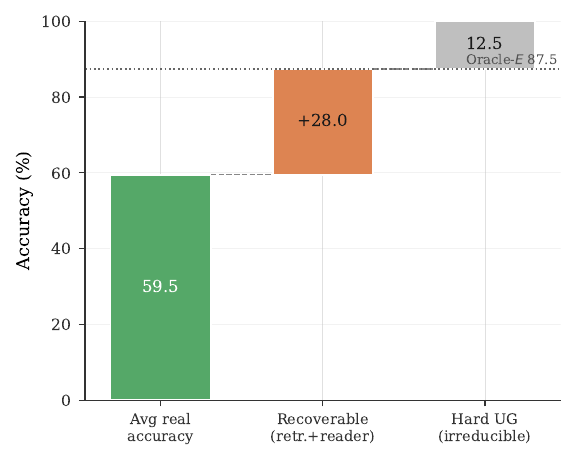}
\caption{Oracle ladder. Real accuracy ($59.5\%$) plus the recoverable gain ($+28.0$\,pp, a joint retrieval$+$reader bound) reaches the Oracle-$E$ ceiling ($87.5\%$).}
\vspace{-4mm}
\label{fig:oracle-ladder}
\end{figure}

\paragraph{What drives the utilization gap?} A UG case means the agent \emph{found} gold evidence yet still answered wrong, either through genuine reasoning failure or benchmark/evaluation artifacts. We probe the distinction in two ways. First, we strip \texttt{search} and \texttt{visit} from the harness and inject $E(q)$ or $G(q)$ directly into a shared GPT-4o reader at temperature $0$. The ceilings are Oracle-$E = 87.5\%$ and Oracle-$G = 87.4\%$, only $0.1$\,pp apart, so evidence-set distractor density does not bind. The remaining \emph{Hard UG} is $12.5\%$ (Figure~\ref{fig:oracle-ladder}): queries even a strong reader fails on with the evidence in front of it. Per-agent Net Gain, Oracle-$E$ minus real accuracy, ranges from $+13.4$\,pp for GLM 5.1 to $+49.5$\,pp for gpt-oss-120b and tracks the qrels-RG ranking with Spearman $\rho > 0.95$. Because this oracle swaps in GPT-4o, Net Gain is a \emph{joint} retrieval$+$reader ceiling, not pure retrieval headroom.

Second, we classify each wrong answer by string similarity to the gold answer using normalized SequenceMatcher with substring containment: \emph{subtle} errors have similarity at least $0.85$ and differ only in capitalization or punctuation; \emph{format} errors fall between $0.6$ and $0.85$ and wrap correct content in extra exposition or markdown; \emph{misunderstanding} errors fall below $0.6$. Kimi K2.6 has the largest format-or-subtle rate among answering agents at $19.9\%$, with GLM 5.1 at $18.6\%$ and Qwen3.5-35B-A3B at $17.7\%$ close behind. This is essentially a free loss: stripping markdown and normalizing case would recover roughly that fraction without touching search or reasoning. Tongyi-DR's near-total misunderstanding rate of $99.0\%$ is instead an artifact of the ``No answer found.'' string emitted by overflow incompletes. UG is thus heterogeneous: part genuine reasoning failure, part systematic evaluation artifact.

\paragraph{Retrieval gap is a query-side, not retriever-side, failure.} Is RG an agent-side query-formulation failure or a retriever-side recall-ceiling failure? All six agents share the same Qwen3-Embedding-8B with top-$K{=}5$ results (§\ref{sec:signals}), yet mean gold recall ranges from gpt-oss-120b's $52.0\%$ to GLM 5.1's $78.7\%$. A $26$-point spread on a fixed retriever can only be query-side. The retriever sets an absolute ceiling that bounds all agents alike, but the variance and per-agent leverage are in query formulation, which §\ref{sec:results-query} takes up in detail.

\subsection{RQ3: Episode-Level Mechanism}
\label{sec:results-mechanism}
\label{sec:results-efficiency}
\label{sec:results-utility}

\emph{What does each trajectory look like turn by turn, and what produces the outcomes above?} We read four local properties: episode utility, productive-episode position, visit selectivity, and correct-vs-incorrect contrast.

\paragraph{$77\text{--}94\%$ of episodes add no new evidence; waste is not uniform across queries.} With $R_k$, $C_k$, and $\Delta_k$ from §\ref{sec:results-attribution} and $k^*(\xi) = \max\{k : \Delta_k(\xi) \neq \emptyset\}$, each episode is \emph{productive} if $\Delta_k \neq \emptyset$, \emph{redundant} if $R_k \neq \emptyset$ but $\Delta_k = \emptyset$, and \emph{unproductive} if $R_k = \emptyset$. Figure~\ref{fig:traj-combined}a shows that only $6\text{--}23\%$ of episodes are productive; $20\text{--}29\%$ are redundant and $50\text{--}70\%$ unproductive, so $77.5\text{--}93.6\%$ add no new evidence. Waste is not uniform: $8\text{--}24\%$ of questions return zero productive episodes, from Kimi K2.6 low to gpt-oss-120b high, while the median productive count is only $2\text{--}3$ everywhere. The result is a zero-productive tail needing reformulation plus a low-productivity bulk needing stopping rules.

\paragraph{Useful evidence concentrates in the early episodes.} Productive episodes concentrate early (Figure~\ref{fig:traj-combined}c,d). By $K \approx 25$, cumulative recall reaches $\mathrm{Recall}_K \in [0.47, 0.70]$, with roughly half already accumulated in the first five episodes. The first-gold-hit step is sharper: by $K = 5$, $28.8\%$ of Tongyi-DR trajectories and $51.9\%$ of GLM 5.1 trajectories have retrieved gold; by $K = 25$, every agent is within $4\text{--}10$\,pp of its gold-hit plateau, $59.6\text{--}86.0\%$, and by $K = 40$ within $1\text{--}5$\,pp. The remaining never hit gold at any budget: what an agent misses early it rarely recovers late. Overall, a trajectory either lands on gold early or not at all; this is why early caps cost little.

\paragraph{Visit precision is per-question bimodal: visits are all-on-target or all-off-target.} Visit precision $\Pr[d \in E(q) \mid (\textsc{visit}, d)]$ is the click-model analogue of search-result quality applied to the agent's own clicks.\footnote{In our closed-corpus regime, visit precision coincides numerically with the visit-level redundant rate; we frame as click precision because that interpretation generalizes to the open-web setting.} It varies sharply across agents (Figure~\ref{fig:acc-vs-x}b; Table~\ref{tab:overall}). Tongyi-DR is the volume-vs-selectivity outlier, opening $3.81$ documents per question at $49.5\%$ precision; the selective frontier cluster opens at most $1.8$ at roughly $80\%$ precision or better. Per question, the distribution is strongly bimodal: for the frontier cluster, $71\text{--}79\%$ of questions achieve $100\%$ precision and $10\text{--}16\%$ land at $0\%$; for Tongyi-DR, only $28.1\%$ hit $100\%$ and $32.5\%$ land at $0\%$. A snippet-level gate would help where the agent clicks blindly, not uniformly.

\begin{figure}[t]
\centering
\includegraphics[width=\linewidth]{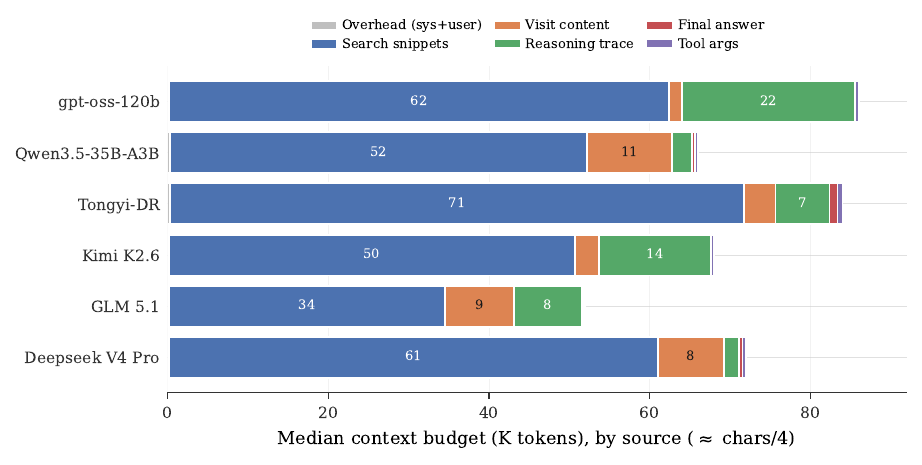}
\caption{Per-agent context budget by source. Bar length is median final context; segments are the per-source split (character-count proxy). Search snippets dominate ($66\text{--}85\%$); visit content is a minority ($2\text{--}16\%$); reasoning fills the rest.}
\vspace{-2mm}
\label{fig:context-composition}
\end{figure}

\begin{table*}[t]
\centering
\small
\caption{Double dissociation between two query styles on two questions where GLM 5.1 and Kimi K2.6 disagree (\textcolor{teal}{$\checkmark$}/\textcolor{red}{$\times$}; \emph{Turns} = search-bearing turns, \emph{Srch} = search calls).}
\label{tab:query-case}
\begin{tabular}{@{}cl cc >{\raggedright\arraybackslash}p{0.55\linewidth}@{}}
\toprule
& Agent & Turns & Srch & Characteristic query move \\
\midrule
\multicolumn{5}{@{}p{0.97\linewidth}@{}}{\emph{Q1 --- name the entity at the end of a constraint chain: a $2020$ article author (fixed publication cadence) interviewed by the co-creator of a Tribeca audio selection} \;(gold \textbf{All Fives})} \\
\addlinespace[2pt]
\textcolor{teal}{$\checkmark$} & GLM 5.1 & 19 & 37 & \emph{Anchor-then-verify}: alternate \texttt{add\_constraint}/\texttt{drop\_constraint} to fix the interviewer, isolate the author (Anjuan Simmons), and converge ($0$ repeats). \\
\textcolor{red}{$\times$} & Kimi K2.6 & 90 & 90 & \emph{Never anchors}: guesses the game and the life skill (``Animal Crossing,'' ``patience,'' \ldots\ across $54$ fresh pivots) and exhausts its budget without the author. \\
\addlinespace[5pt]
\multicolumn{5}{@{}p{0.97\linewidth}@{}}{\emph{Q2 --- identify a US state from four clues ($5$th-tallest capital building, $2.1\%$ population growth, governor's marriage year, borders a neighboring-country region), then name its state bird} \;(gold \textbf{Mountain Bluebird} --- the state is Idaho)} \\
\addlinespace[2pt]
\textcolor{teal}{$\checkmark$} & Kimi K2.6 & 37 & 37 & \emph{Enumerate-then-check}: sweep candidate states and capitals (Austin, Olympia, Sacramento, Phoenix, \ldots), test each clue, converge on Idaho $\to$ Mountain Bluebird. \\
\textcolor{red}{$\times$} & GLM 5.1 & 20 & 38 & \emph{Anchors a misreading}: reads ``borders a region of a neighboring country'' as licensing a foreign answer, commits to Uttarakhand, India, rationalizes every remaining clue to fit, and reports Himalayan Monal. \\
\bottomrule
\end{tabular}
\end{table*}

\paragraph{Incorrect trajectories are $2\times$ longer than correct ones but saturate at similar $k^*$.} Aggregate median $k^*$ is tight across agents, from $3$ for Qwen3.5-35B-A3B to $9$ for Tongyi-DR and Kimi K2.6, but median wasted tail $M-k^*$ diverges from $1$ episode for GLM 5.1 to $19$ for gpt-oss-120b; Tongyi-DR's mean wasted tail of $15.0$ is the proximate driver of its overflow. Splitting by correctness (Figure~\ref{fig:traj-combined}b) sharpens this: incorrect trajectories are $1.9\text{--}2.9\times$ longer than correct ones in mean $M$, yet their $k^*$ arrives only $0.6\text{--}6.7$ episodes later. The real gap is the wasted tail, $3.0\text{--}6.1\times$ longer on incorrect trajectories. Gold recall on incorrect trajectories drops from roughly $92\%$ on the correct set to $27\text{--}45\%$, and $48\text{--}64\%$ of them never surface a gold-bearing document at all. The long wasted tail is therefore \emph{rational continuation under failed retrieval}, not simply a stopping failure: the agent keeps searching because it lacks the answer, but its queries cannot surface gold at any budget. The prescription splits: \emph{query reformulation triggered by saturation-without-gold} for the RG majority, and saturation stopping for the UG subset, where post-gold redundancy of $58\text{--}74\%$ bounds the reclaim.

\paragraph{Accumulated search snippets, not visited documents, dominate the context budget.} The wasted tail is paid for in context tokens, so we ask \emph{which} text fills the window. Bucketing every message by source shows search snippets occupying $66\text{--}85\%$ of context, visit content only $2\text{--}16\%$, and reasoning $3\text{--}25\%$ (Figure~\ref{fig:context-composition}). Tongyi-DR is the extreme at $85.0\%$ snippets: its overflow is the cumulative cost of many top-$K$ searches, not the documents it opens. Visit gating reclaims precision but little budget; the bigger lever is upstream snippet management, including deduplication, top-$K$ trimming, and aging out earlier-episode snippets whose evidence has already been integrated. The pattern reproduces on the open web (§\ref{sec:results-openweb}), so it is not a fixed-corpus artifact.

\subsection{RQ4: Query Strategy Analysis}
\label{sec:results-query}

\emph{The mechanism so far is about how much agents search and where the evidence lands; does the way they query separate strong from weak?} For every search step, we classify the reformulation move by token overlap with the previous query: \texttt{repeat} for Jaccard overlap at least $0.85$, \texttt{pivot} below $0.25$, and \texttt{add}/\texttt{drop\_constraint} in between. We read these moves with the redundancy counters in Table~\ref{tab:query-discipline} and correlate each with accuracy using Spearman $\rho$ over six agents, descriptively here and formalized within-question in §\ref{sec:results-synthesis}. The headline: \emph{effort does not explain the gap}. The weakest agent, gpt-oss-120b at $38.0\%$, issues the most searches, $31.8$ search-turns, and volume runs the wrong way for an effort story, anti-correlated with accuracy at $\rho = -0.77$; what tracks accuracy is query \emph{cleanliness} and \emph{yield}.

\paragraph{Redundant re-querying is the cleanest behavioral predictor of failure.} gpt-oss-120b is the one genuine offender: it re-issues near-identical consecutive queries on $5.0\%$ of moves, while the other five stay at or below $1.5\%$ \texttt{repeat}. Across agents, \texttt{repeat} rate is the strongest non-mechanical behavioral correlate of accuracy, with $\rho = -0.83$. Tongyi-DR's raw \texttt{repeat} rate of $4.8\%$ looks like a second offender, but it is an artifact of malformed calls: its scaffold passes \texttt{query} as a JSON \emph{array}, which the harness's single-string schema rejects. That mismatch, not empty retrieval, explains its $11.3\%$ \emph{Bad-call} rate; excluding rejected calls drops genuine \texttt{repeat} to $1.5\%$, squarely in the clean range reported in Table~\ref{tab:query-discipline}. We read \emph{Bad-call} as a harness--scaffold mismatch rather than a retrieval-strategy signal, with the caveat that roughly one in nine intended Tongyi-DR searches never reaches the retriever; its $38\%$ overflow is instead driven by post-gold redundancy and snippet volume (§\ref{sec:results-mechanism}).

\paragraph{Fewer reasoning hops, via parallel query batching --- but this is a style, not a law.} The strongest agent is also the most economical: GLM 5.1 resolves a query in $10.6$ search-turns and Qwen3.5 in $12.4$, compared with $27\text{--}32$ for gpt-oss-120b, Tongyi-DR, and Kimi K2.6. The lever is batching: GLM 5.1 and Qwen3.5 fire $1.87$ and $1.74$ queries \emph{per turn}, trying several angles before reacting once, whereas the other three issue about one query per turn. Batching correlates positively with accuracy at $\rho = +0.75$. Still, turn economy is a style, not a requirement: Kimi K2.6 reaches $69.2\%$ over $27$ serial turns, brute-forcing breadth where GLM 5.1 triangulates. This is why raw search-turn count remains a confounded predictor.

\begin{table}[t]
\centering
\small
\setlength{\tabcolsep}{3pt}
\caption{Per-agent query-strategy profile on BrowseComp-Plus.  \emph{Q/turn}: queries per search turn ($>1$ = parallel batching); \emph{Pivot}: share of moves that jump to a new angle; \emph{Repeat}: share of near-identical consecutive queries; \emph{Bad-call}: share of malformed search calls.}
\label{tab:query-discipline}
\begin{tabular}{lcccccc}
\toprule
        & Acc  & Search  & Q/    & Pivot & Repeat & Bad  \\
Agent   & (\%) & turns   & turn  & (\%)  & (\%)   & (\%) \\
\midrule
gpt-oss-120b           & 38.0 & 31.8 & 1.00 & 46.6 & 5.0 &  0.3 \\
Tongyi-DR              & 52.2 & 27.7 & 1.00 & 58.3 & 1.5 & 11.3 \\
Qwen3.5-35B-A3B        & 55.2 & 12.4 & 1.74 & 44.1 & 0.4 &  0.0 \\
Deepseek V4 Pro        & 68.6 & 19.3 & 1.49 & 55.9 & 0.3 &  0.0 \\
Kimi K2.6              & 69.2 & 27.1 & 1.01 & 70.1 & 0.8 &  0.0 \\
GLM 5.1                & 74.1 & 10.6 & 1.87 & 47.5 & 0.0 &  0.0 \\
\bottomrule
\end{tabular}
\end{table}

\paragraph{Trying new angles is not the waste --- failing to \emph{land} them is.} One might expect weaker agents to thrash through more dead-end pivots. They do not: \texttt{pivot} rate is not negatively correlated with accuracy, with $\rho = +0.31$, and the most pivot-heavy agent, Kimi K2.6 at $70\%$, is among the strongest. What separates strong from weak is whether new angles \emph{land} on evidence. Gold coverage broadly tracks accuracy, from $59.6\%$ to $86.0\%$ (§\ref{sec:results-mechanism}), even though pivot rates do not order the agents. gpt-oss-120b spends the most search-turns and pivots on nearly half its moves, yet converts that motion into gold least often.

\paragraph{Case studies: neither query style dominates.} The two styles are agent-level tendencies visible in Table~\ref{tab:query-discipline}: Kimi K2.6 pivots the most, at $70\%$ of moves, and surfaces the most unique documents in its search results, $60.5$ per question, giving it a breadth-first style, whereas GLM 5.1 commits faster by anchoring and verifying a candidate entity. Neither style is uniformly better. Across the $830$ questions, GLM 5.1 is correct where Kimi K2.6 fails on $109$, and Kimi K2.6 is correct where GLM 5.1 fails on $68$. Table~\ref{tab:query-case} reads one case in each direction. Anchoring wins when the right entity is anchorable: in Q1, GLM 5.1 isolates the author in $19$ turns while Kimi K2.6 burns $90$ without converging. It \emph{backfires} when the model commits to a misread clue: in Q2, GLM 5.1 reports a Himalayan bird for what is actually Idaho, while Kimi K2.6's breadth-first sweep lands the right answer with a near-identical search count, $38$ versus $37$. Each style has a failure mode, so which pays off is set by the question, not model strength alone.

\subsection{RQ5: Why Strong Agents Win}
\label{sec:results-synthesis}

\emph{Do the recall and query-discipline effects hold within the same question, or do they merely proxy generic model capability?} The cross-agent correlations of §\ref{sec:results-attribution} and §\ref{sec:results-query} are descriptive over six points. To separate capability from behavior, we pool the $830$ questions into an agent$\times$question panel with $4{,}919$ usable trajectories after dropping too-short runs, then fit linear probability models with \emph{question fixed effects}: every comparison is between agents answering the same question. Predictors are $z$-scored and include malformed-call-corrected \texttt{repeat}/\texttt{pivot} shares, queries per turn, and log search calls; standard errors cluster by question.

\begin{figure}[t]
\centering
\includegraphics[width=\linewidth]{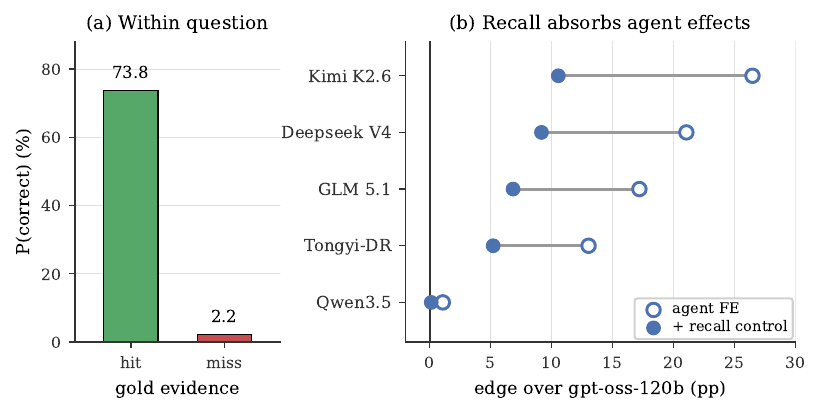}
\caption{Within-question test: \emph{(a)} P(correct) on the $366$ gold-hit-disagreement questions; \emph{(b)} agent edge over gpt-oss-120b (question FE) before (open) / after (filled) recall control --- $56\text{--}60\%$ absorbed.}
\vspace{-2mm}
\label{fig:panel-mediation}
\end{figure}

\paragraph{Retrieval mediates the outcome, within the question.} On the $366$ questions where agents disagree about whether gold evidence was retrieved, the retrieving agents answer correctly $73.8\%$ of the time, while the others answer correctly only $2.2\%$ on those same questions (Figure~\ref{fig:panel-mediation}a). The near-zero $2.2\%$ bounds closed-book answering: agents almost never succeed without retrieving gold, so recall is not merely a proxy for parametric knowledge, though it cannot separate query skill from knowledge-informed querying. In the regression, cumulative gold recall dominates with a $+30$\,pp effect per SD and $t = 34$, an order of magnitude above every behavioral coefficient. Retrieval also carries the bulk of the \emph{identity} advantage: controlling for recall absorbs $56\text{--}60\%$ of each stronger agent's conditional edge over gpt-oss-120b in Figure~\ref{fig:panel-mediation}b.

\paragraph{Query discipline acts through retrieval; search volume is a symptom.} The behavioral signature of §\ref{sec:results-query} decomposes cleanly on the panel. \texttt{repeat} predicts retrieval failure within the question, reducing gold-hit probability by $1.9$\,pp per SD with $t = -3.4$; its association with correctness vanishes once recall is controlled, and it loses significance under agent fixed effects. Search \emph{volume} is strongly negative within the question: where agents disagree, the correct ones issue on average $14$ fewer search calls while achieving $54$\,pp higher gold recall, and the coefficient survives restricting to answered-only trajectories. We read this associationally: an agent searches long \emph{because} it is failing at least as much as the reverse. \texttt{pivot} is null throughout; per-turn batching is the one behavior whose positive effect survives agent fixed effects and recall control, adding $+2\text{--}4$\,pp per SD.

\subsection{RQ6: External Validity on Open Web}
\label{sec:results-openweb}

\emph{Is the coverage--effort dissociation an artifact of the fixed corpus and qrels supervision?} We re-run the same ReAct harness on a different benchmark and retrieval regime: a $200$-question sample of BrowseComp-en \citep{Wei2025BrowseCompASimple} answered against a live commercial open-web search API, using the four agents that ran to completion within budget. Because the open web has no qrels, an \emph{LLM judge} supplies the relevance signal: for each question, it decomposes the information need, labels whether every \texttt{search}/\texttt{visit} surfaces a \emph{critical new fact}, and judges whether the gathered facts are jointly \emph{sufficient}. \emph{Coverage} is the share of questions judged sufficient; an error is a \emph{retrieval gap} when sufficiency is never reached and a \emph{utilization gap} otherwise. Answer accuracy keeps the same GPT-4o protocol.

\paragraph{The headline replicates.} Across the four agents run to completion on the open web, accuracy tracks judged coverage and is flat against effort. GLM 5.1 has the top accuracy at $64.0\%$ with the fewest calls, about $65$ per question; Deepseek V4 Pro spends the most, $81$ calls per question, for slightly lower accuracy at $61.5\%$; and the two mid-scale agents, Tongyi-DR and Qwen3.5-35B-A3B, remain below $45\%$, at $42.5\%$ and $30.5\%$. With only four agents, the correlations are directional, but accuracy--coverage has $r = 0.99$ while accuracy--calls has $r = 0.12$. As in the main study, retrieval gap dominates errors, accounting for $68\text{--}88\%$ of them. The reformulation move-share is identical across the correct/incorrect split, so what separates outcomes is whether a trajectory reaches \emph{sufficient evidence}, not how it queries. The snippet-dominated budget also reproduces.

\paragraph{The qrels-free utility axes behave consistently across benchmarks.} The episode-utility decomposition of §\ref{sec:results-mechanism} can be reconstructed without qrels because the LLM usefulness judge labels each episode, search step, and visit as \emph{productive}, \emph{redundant}, or \emph{unproductive}. On BrowseComp-Plus, productive$+$redundant share correlates with accuracy most strongly at the \emph{episode} level, where $r = 0.71$, weakening through search steps at $r = 0.63$ to visits at $r = 0.51$. This is the qrels-free echo of recall-over-precision (§\ref{sec:results-attribution}). The four-agent open-web check in Figure~\ref{fig:axes-compare} is directionally consistent but underpowered for the ordering, so we lean on BrowseComp-Plus for that.

\ifcompact
\begin{figure}[t]
\centering
\includegraphics[width=\linewidth]{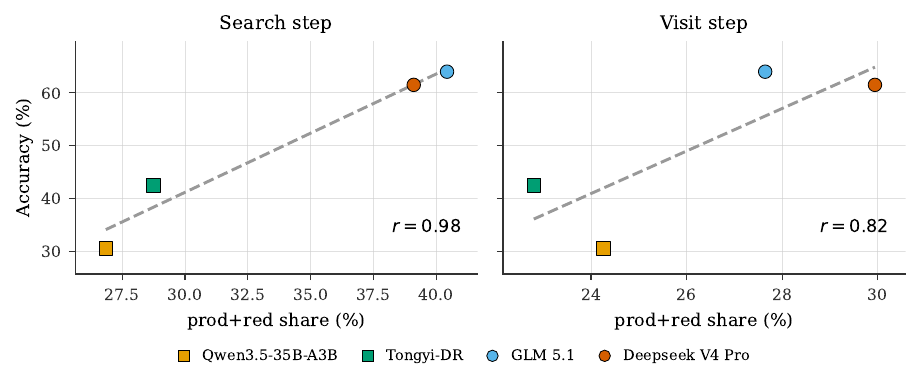}
\caption{Accuracy vs.\ productive$+$redundant share of search / visit steps on BrowseComp-en.}
\label{fig:axes-compare}
\end{figure}
\fi

\section{Discussion}
\label{sec:discussion}

\paragraph{What makes an agent strong.} Every layer returns the same answer: strong agents win by writing queries that retrieve, and nearly everything else is downstream. Accuracy is dominated by retrieval recall rather than context budget; within the same question, the agent that retrieves the gold evidence usually answers correctly, while the one that misses it almost never does. Recall also absorbs most of the apparent between-agent capability gap (§\ref{sec:results-synthesis}). The behavioral signature of strength is discipline, not effort: strong agents essentially never re-ask the same content, pivot as often as weak agents but \emph{land} those pivots, often batch several angles per turn, and stop near evidence saturation. High search volume is therefore a symptom of a failing trajectory, not a lever that rescues it.

\paragraph{What transfers to weaker agents, and how.} Similar accuracies can hide disjoint failure profiles: Kimi K2.6 is utilization-bound while Deepseek V4 Pro is retrieval-bound at the same accuracy, and Tongyi-DR's apparent search problem is largely a missing stopping rule. Transfer should therefore be prescribed by diagnosis, not leaderboard position. The largest transferable surface is the query side. On the shared retriever the $26$-point spread in gold recall is query-side by construction (§\ref{sec:results-failure}), so retrieval-gap errors are not benchmark-impossible but query-fixable in principle. Possible mechanisms include reformulation and deduplication guards \citep{huang2009analyzing}, prompt-level strategy transfer, or distillation on stronger agents' queries, though our numbers are upper bounds rather than realized gains. Other levers are scaffold-side and cheap: answer normalization for format-or-subtle errors, schema repair for rejected searches, snippet-stream management, and snippet-first visit gates \citep{Chen2026AgentIRReasoning-AwareRetrieval}.

\paragraph{What remains at the frontier.} For the strongest agents the diagnosis shifts. GLM 5.1 sits closest to the oracle reader ceiling with a negligible wasted tail, so reformulation triggers and stopping rules have little left to reclaim. Its residual errors concentrate in mis-anchoring (Q2 of Table~\ref{tab:query-case}) and in the format-or-subtle share, where the prescriptions are verification and answer normalization rather than more search. This suggests a different frontier problem: not how to spend more retrieval budget, but how to decide when an apparently coherent chain is anchored to the wrong entity, and how to normalize answers without weakening precision. Complementarity also matters: agents fail on largely disjoint questions, so no single agent's error set marks the field's ceiling. The oracle reader places that ceiling at $87.5\%$ (§\ref{sec:results-failure}), leaving a residual that gold evidence in hand does not fix.

\paragraph{Relation to human search behavior.} The patterns also clarify where current agents differ from human searchers. Berrypicking \citep{bates1989berrypicking} predicts that information needs evolve as evidence accumulates; our agents often do the opposite, fixing an initial framing and then issuing paraphrase searches long after new evidence has stopped arriving \citep{huang2009analyzing, rieh2006analysis}. Humans also stop under satisficing, fatigue, and the economics of search cost \citep{zach2005enough, azzopardi2014economic}; agents usually stop at hard caps or after emitting an answer. The missing piece is not merely a shorter budget, but an internal sense of marginal utility: whether a new query is likely to change the answer, whether a snippet adds new evidence, and whether enough independent support has accumulated. In that sense, evidence-driven stopping and query reformulation are not just engineering tricks; they are the agentic analogues of human search judgment.

\section{Conclusion}
\label{sec:conclusion}

We studied long-horizon search agents by opening the search trajectory rather than treating it as an opaque path to a final answer. Across six agents on BrowseComp-Plus, under a shared ReAct harness and retriever, effort and outcome dissociate: neither search volume nor context budget explains accuracy. What does is whether an agent's queries retrieve the gold evidence, and the behavioral signature of the agents that do is query discipline. The trajectory analysis shows why: useful evidence arrives early or not at all, while later turns run on as a wasted tail of redundant snippets. The failure analysis splits errors into retrieval gaps, where evidence was never surfaced, and utilization gaps, where it was surfaced but not converted into the right answer; a within-question test confirms that this is not merely a capability ranking --- on the same questions, retrieving the gold evidence is what separates success from failure.

These findings argue for making search better directed rather than deeper: for weaker agents the main opportunity is query formulation, recognizing failed hypotheses and avoiding redundant re-querying; for stronger agents it shifts toward verification and answer normalization. Our analysis quantifies the headroom but remains diagnostic: realizing the gains will require training, prompting, or tool-design changes.

\bibliographystyle{ACM-Reference-Format}
\bibliography{ref, search_agent}


\end{document}